\pdfoutput=1

\documentclass[11pt]{article}

\usepackage[preprint]{acl}

\usepackage{times}
\usepackage{latexsym}

\usepackage[T1]{fontenc}

\usepackage[utf8]{inputenc}

\usepackage{microtype}

\usepackage{inconsolata}

\usepackage{graphicx}

\usepackage{multirow}
\usepackage{array}
\usepackage{booktabs}
\usepackage{float}
\usepackage[table]{xcolor}
\usepackage{listings}
\usepackage{times}
\usepackage{latexsym}

\usepackage[T1]{fontenc}

\usepackage[utf8]{inputenc}

\usepackage{microtype}

\usepackage{inconsolata}

\usepackage{listings}
\usepackage{graphicx}

\usepackage{caption}
\usepackage{subcaption}
\usepackage{enumitem}

\usepackage{todonotes}
\newcounter{todocounter}

\title{Argument-Based Comparative Question Answering\\ Evaluation Benchmark }

\author{
 \textbf{Irina Nikishina\textsuperscript{1}},
 \textbf{Saba Anwar\textsuperscript{1}},
 \textbf{Nikolay Dolgov\textsuperscript{2}}, \\
 \textbf{Maria Manina\textsuperscript{2}},
 \textbf{Daria Ignatenko\textsuperscript{2}},
  \textbf{Viktor Moskvoretskii\textsuperscript{2,4}},
 \\
  \textbf{Artem Shelmanov\textsuperscript{3}},
  \textbf{Tim Baldwin\textsuperscript{3}},
 \textbf{Chris Biemann\textsuperscript{1}}
\\
\\
 \textsuperscript{1}University of Hamburg,
 \textsuperscript{2}HSE University,
 \textsuperscript{3}MBZUAI,
 \textsuperscript{4}Skoltech
\\
 \small{
   \textbf{Correspondence:} \href{mailto:irina.nikishina@uni-hamburg.de}{irina.nikishina@uni-hamburg.de}
 }
}

\begin{document}
\maketitle
\begin{abstract}


In this paper, we aim to solve the problems standing in the way of automatic comparative question answering. To this end, we propose an evaluation framework to assess the quality of comparative question answering summaries. We formulate 15 criteria for assessing comparative answers created using manual annotation and annotation from 6 large language models and two comparative question asnwering datasets. We perform our tests using several LLMs and manual annotation under different settings and demonstrate the constituency of both evaluations. 
Our results demonstrate that the Llama-3 70B Instruct model demonstrates the best results for summary evaluation, while GPT-4 is the best for answering comparative questions. All used data, code, and evaluation results are publicly available\footnote{\url{https://anonymous.4open.science/r/cqa-evaluation-benchmark-4561/README.md}}.
\end{abstract}


\section{Introduction}

Are \textit{``Nike''} sneakers better than \textit{``Adidas''}? Is an \textit{``iPhone''} easier to use than an \textit{``Android''} phone? What is faster, \textit{``Windows''} or \textit{``Linux''}? Such questions are common user requests on the web. Making an informed choice requires evaluating and comparing various characteristics of the options under consideration. For instance, when buying a car, it is important to consider the brand reputation, the car's fuel efficiency, comfort, and general performance. Conducting thorough comparisons manually by scanning different websites requires time and effort, especially given the vast amount of data available online~\citep{iso-etal-2022-comparative}. In such cases, a comparative answer or summary 
can serve as an effective and time-saving solution.

\begin{figure}[ht!]
    \centering
    \includegraphics[width=\linewidth]{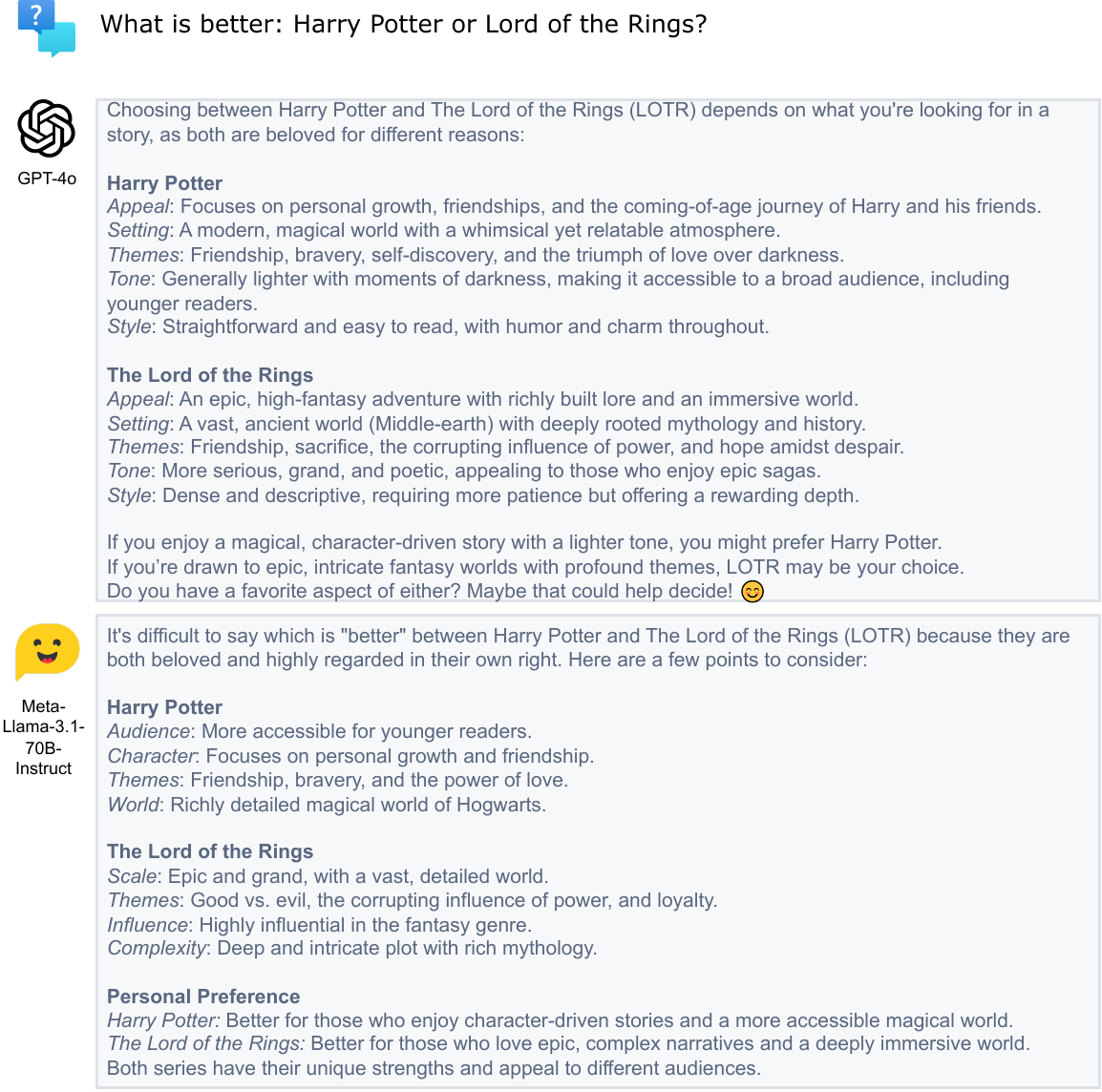}
    \caption{Example of a LLM answers to comparative questions (shortened).}
    \label{fig:example_intro}
\end{figure}

\begin{figure*}
    \centering
    \includegraphics[width=\linewidth]{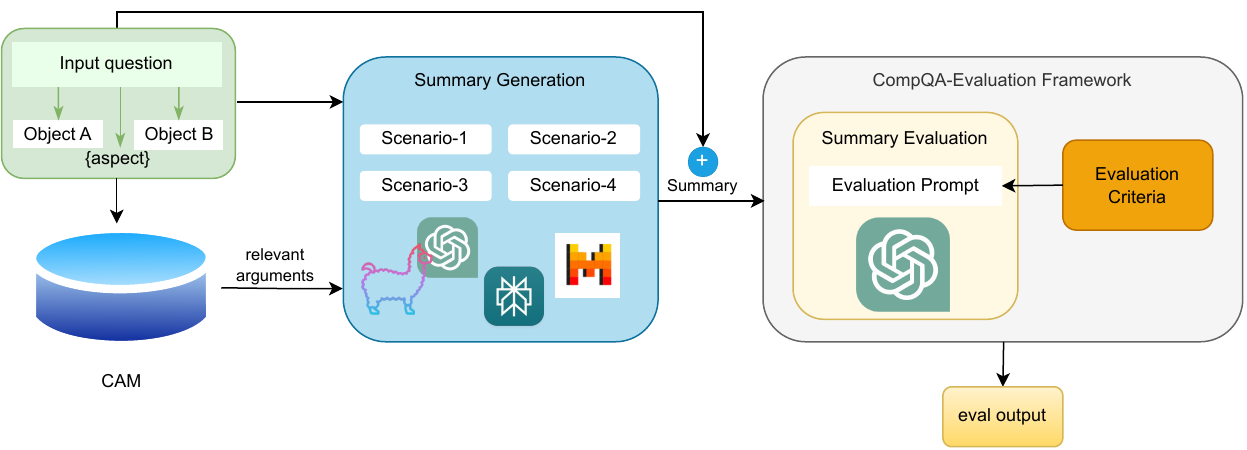}
    \caption{Overall pipeline of the CompQA evaluation dataset creation and evaluation framework.}
    \label{fig:pipeline}
\end{figure*}

Comparative Question Answering (CQA) is a Natural Language Processing task that aims to answer requests containing two or more objects by choosing one of them and supporting the answer with arguments. The final goal is not necessarily to come to a distinct final choice, but to describe each object's advantages and disadvantages, taking into account the specified aspect(s). This task can also be viewed as an abstractive comparative summarization.
In this paper, we focus on comparing only two objects, leaving the cases with three or more for future research.
Unlike traditional opinion summarization, which focuses on aggregating views about a single object, the task of generating comparative summaries is more complex, as it involves evaluating each object in the context of the other. 
Traditional approaches to this task 
focus only on the initial steps of CQA that precede comparative answer generation 
and only partially address the Argument Summarization task \cite{chekalina-etal-2021-better,shallouf-etal-2024-cam} for answer generation.

Large Language Models (LLMs) have emerged as powerful tools for a wide range of tasks, including search and recommendations. They significantly simplify the process of generating comparative answers, often delivering coherent and well-structured responses (see example in Figure~\ref{fig:example_intro}). However, the quality of these responses varies greatly, posing a persistent challenge in the automated evaluation of such summaries. 
As agreed upon in previous work, a well-constructed abstractive summary should be \textit{coherent}, \textit{concise}, \textit{factually consistent}, \textit{relevant}, \textit{non-redundant}, \textit{grammatically accurate}, and of \textit{decent readability}~\citep{GUPTA201949, SHAKIL2024128255}. Achieving this type of qualitative assessment often requires human evaluation, guided by a structured framework that breaks the evaluation into specific criteria. Importantly, this framework may vary depending on the nature of the summary: for instance, summarizing a Wikipedia article demands different criteria (e.g. factual sequences) as opposed to summarizing a comparison between two products. Thus, no consensus regarding the actual criteria a summary must follow. Moreover, such manual annotation is a tedious task, so its automation is highly beneficial for the research and development of CQA systems.

In this paper, we aim to fill the mentioned gaps in benchmarking the CQA systems and answer the following research questions:
\textit{(RQ1) Which criteria should be considered for assessing a comparative summary?}, \textit{(RQ2) Can LLMs reliably evaluate comparative summaries with human expert-level quality?}, \textit{(RQ3) How do different LLMs fare against each other in generating high-quality summaries?} 

The contributions of this work are as follows:

\begin{enumerate}[itemsep=0pt, topsep=0pt]

\item Systemizing previous efforts in CQA, we 
develop 15 criteria for scoring comparative answers and implement automatic CompQA evaluation pipeline based on a LLM-as-a-judge approach (Figure~\ref{fig:pipeline}) that is able to assess CQA summaries on the basis of these criteria. We verify the usefulness of the proposed benchmark by comparing automatic assessments with human judgements.

\item Using the developed benchmark, we perform the first automatic and manual evaluation of 6 contemporary LLMs on the CQA task (GPT-3.5, GPT-4, Llama3-8B Instruct, Llama3-70B Instruct, Perplexity, and Mixtral), the CAM 2.0 summary dataset \cite{shallouf-etal-2024-cam}, and Yahoo.Answers from the dataset of \citet{chekalina-etal-2021-better}. 

\item We create a dataset of comparative questions with object pairs, aspects, and arguments for the CQA text generation task along with manually corrected answers. 


\end{enumerate}

All code, datasets, and evaluation results are available online.\footnote{\url{https://anonymous.4open.science/r/cqa-evaluation-benchmark-4561/README.md}}

\section{CompQA Evaluation Framework} \label{sec:eval_framework}
Defining the criteria for a high-quality answer is crucial, as just checking for coherence and logic, or metrics like word and vector similarity is not enough. This is especially relevant to our use-case, as, when generating answers, we use a list of arguments, and the output is directly tied to the complexity and level of detail of the input prompt~\citep{loya-etal-2023-exploring}. Thus, the generated result can vary from an unstructured paragraph to a well-organized summary with clearly defined components, depending entirely on the input.
This also means that two summaries comparing the same pair of objects can and will have a low token match metric if different arguments were used, despite the fact that they may both still be high-quality comparative texts.

Considering two objects are to be compared, we define a good-quality comparative answer as one that helps the user \textit{decide} between the objects by comparing several \textit{relevant aspects}. These aspects can be general or specified by the user. In developing our idea of such a prompt, we assume that a summary should be \textit{well-structured}, \textit{concise}, \textit{factual}, \textit{relevant}, \textit{coherent}, \textit{informative}, and \textit{non-redundant}. 
Thus, to answer the \textbf{RQ-1} on developing criteria for the comparative question-answer summary, we define and categorize the following:
\begin{figure*}[ht!]
    \centering
    \includegraphics[width=0.98\linewidth]{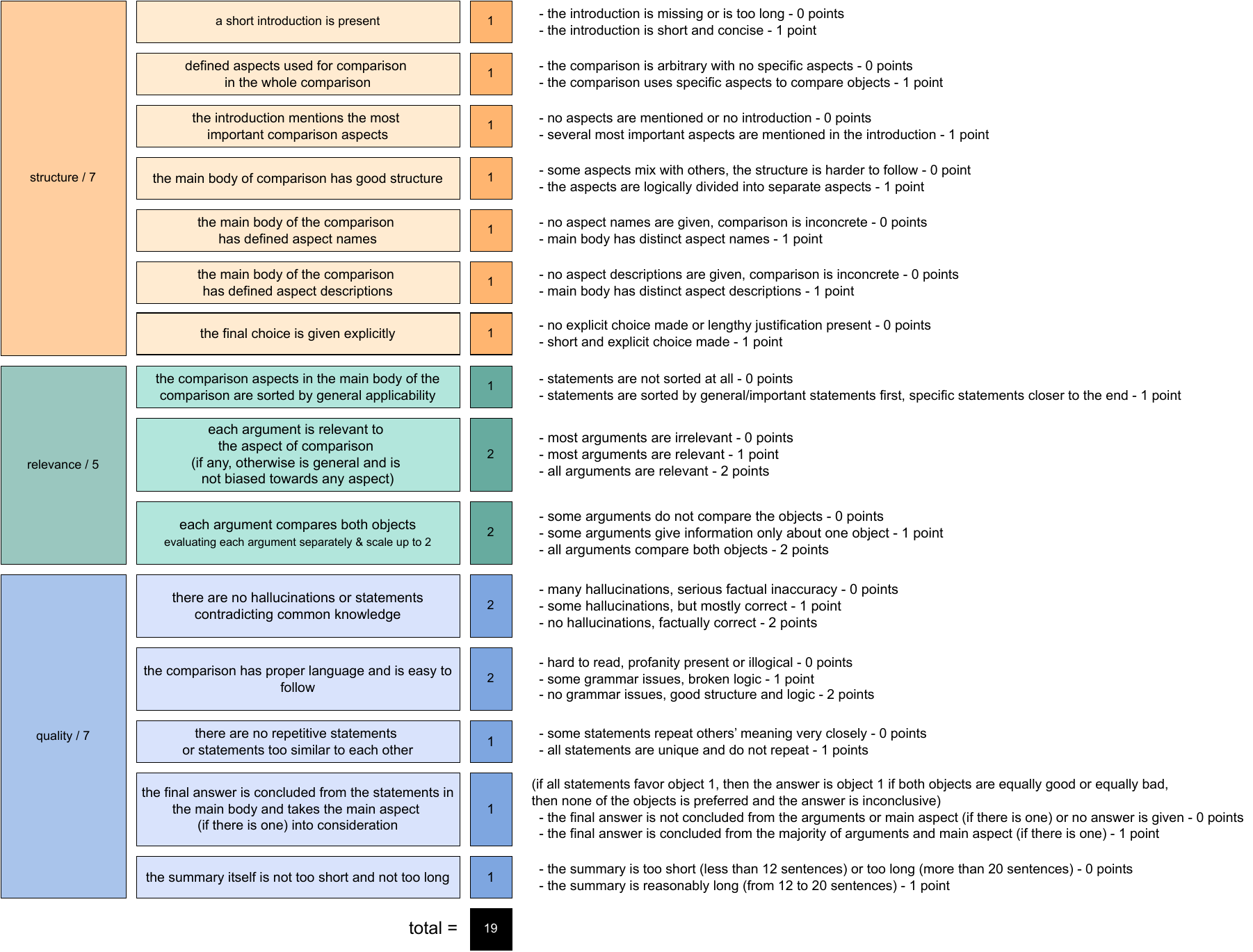}
    \caption{CompQA evaluation framework criteria.}
    \label{fig:framework}
\end{figure*}

\begin{enumerate}[itemsep=0pt, topsep=0pt]
\item \textit{structure}: a well-defined structure that has: a) a short introductory summary, b) a list of \textit{named aspects} with \textit{short descriptions}, and c) a distinct \textit{final choice}. 

\item \textit{relevancy}: arguments are relevant to the aspect of comparison (if given), compare both objects, and are ordered from more generic to specific ones or vice versa. 

\item \textit{quality}: should further fulfill the following aspects: \textit{concise} (optimal length focusing on key information, should be between 12 to 20 sentences), \textit{factual} (no hallucination or contradictory arguments), \textit{informative} (if the summary is required for a specific aspect, then the introduction, arguments, and conclusion should incorporate that, otherwise they should be generic), \textit{coherence} (comparison logic is easy to follow, statements are not self-contradictory or repetitive, the conclusion does not contradict the arguments).
\end{enumerate}

\subsection{Automatic Evaluation using LLM-as-a-Judge}

Based on the described criteria, we present an evaluation framework called CompQA that implements these quality checks with 15 questions and assigns a score between 0 and 19 points. Each question assigns a maximum score of 1 or 2 depending on various scenarios. The criteria are developed to simplify the annotation and increase inter-annotation agreement of human annotators. Figure \ref{fig:framework} presents the CompQA framework and explains each scored point for each criterion in detail.

To automatically score the CQA answers, we prompt an LLM.
The template is presented in Figure \ref{fig:eval_prompt}. The complete prompt scenario can be found in Figures \ref{fig:full_eval_prompt} and \ref{fig:full_eval_prompt2} in Appendix \ref{appendix:eval}. The output of the model is expected to be a JSON dictionary where the keys are criteria numbers and values are the scores assigned by LLM. After preliminary experiments, we have decided not to require models to provide the total score, as the total value is usually not the same as the actual sum. The model provides the scores according to each criteria.



\begin{table}[t]
  \centering
  \small

    \begin{tabular}{l|r}
    \toprule
    \multicolumn{2}{c}{\textbf{Characterstics}} \\
    \midrule
    Total  Pairs  & 50 \\
    Pairs with defined aspect & 20 \\
    Average \# of arguments & 10 \\
    Min \# of arguments & 3 \\
    Max \# of arguments & 20 \\
    Pairs with more than average \# of arguments & 23 \\
    \bottomrule
    \end{tabular}%
     \caption{Statistics of data retrieved from \cite{chekalina-etal-2021-better} and CAM 2.0 \cite{shallouf-etal-2024-cam}.}
  \label{tab:cam_stats}%
\end{table}%

\subsection{Question and Argument Collection}

We also construct a novel dataset of questions and arguments for our benchmark.
First, we randomly select 50 pairs from the Touch{\'e} dataset \cite{bondarenko:2022e} and \cite{chekalina-etal-2021-better} with aspects  and retrieve a maximum of 10 arguments for each object from the CAM 2.0 system \cite{shallouf-etal-2024-cam}. More information about these datasets can be found in Appendix \ref{appendix:data}. The exact number varies depending on factors like the popularity of the object, availability of arguments in its favor on online forums, etc. The pairs from both sources belong to various domains, e.g. companies (\textit{``IBM vs Sony''}), places (\textit{``Virginia vs. Michigan''}), and products of different categories (\textit{``PS3 vs. DS''} and \textit{``tea vs coffee''}), etc. Table~\ref{tab:cam_stats} provides a few key details about this dataset. On average, each pair contains 10 arguments, with the number ranging from 3 to 20. Only 20 pairs are provided a defined aspect as input. If available, we also extract the questions for the pairs; otherwise, a basic question is used as a default: \textit{``What is better: \{object1\} or \{object2\}? Focus on \{aspect\}.''}

\subsection{Prompt Scenarios for Comparative Answer Generation}

We use several prompts with varying levels of complexity and specificity to get summaries from the LLMs for the following reasons:

\begin{itemize}[itemsep=0pt, topsep=0pt]
    \item check whether our framework can differentiate between good and bad summaries;
    \item to assess the capability of LLMs to produce comparative answers of good quality with and without provided arguments;
    \item how the quality of the output answer can vary with prompt-engineering for the same model.

\end{itemize}

\begin{figure}[ht!]
\centering
\includegraphics[width=0.92\linewidth]{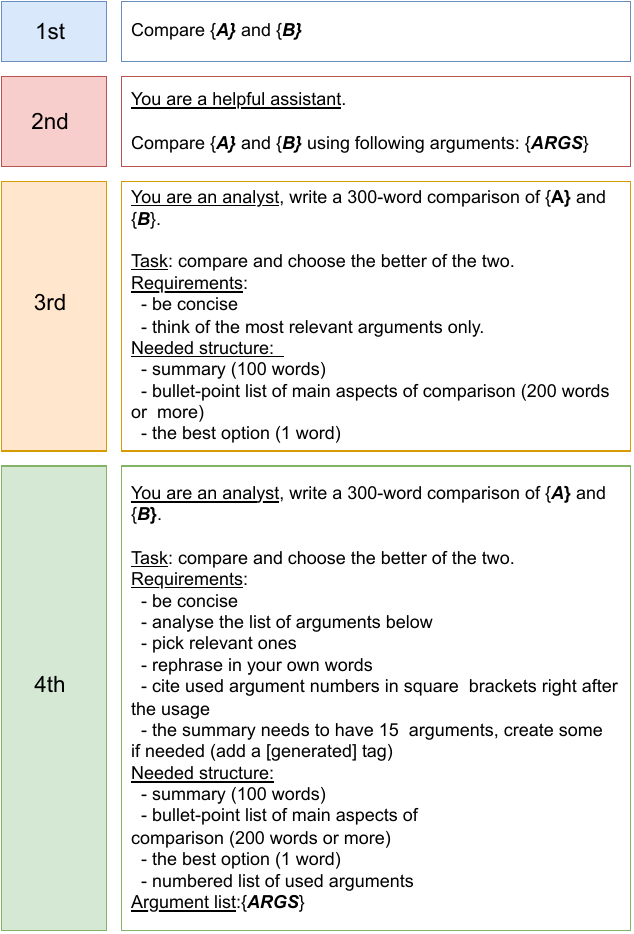}
\caption{Four prompting scenarios used for comparative summary generation. Objects ``A'' and ``B'' are extracted from the input question, ``ARGS'' is the list of arguments taken from CAM.}
\label{fig:prompt_variants}
\end{figure}

Figure~\ref{fig:prompt_variants} shows the exact content of these scenarios.
The \textbf{first} scenario is the simplest one that does not include any details from the user other than the objects to compare. It should produce a summary with an arbitrary structure. 
It is important to note that the remaining three scenarios assign roles to the LLM, whereas the first one does not.

The \textbf{second} scenario gives the LLM a list of arguments (ARGS) extracted from CAM, to see if the LLM uses the given arguments exclusively or generates new ones.

The \textbf{third} and \textbf{fourth} scenarios add more instructions to guide the LLM in producing a summary with a specific structure as described in Section~\ref{sec:eval_framework}. The \textbf{third} scenario removes CAM arguments, testing the LLM's ability to generate its own. 

Lastly, the \textbf{fourth} scenario includes CAM arguments and improves upon the specific instructions. Due to their increased complexity, the third and fourth scenarios aim to produce a summary that should score better than other scenarios.

\section{Evaluation of CompQA via Human Annotation}

In this section, we introduce the models that we have chosen to be potential LLM cores for our agent. To choose the best fitting model, we compare them using the CompQA benchmark, different LLMs as evaluators, as well as human evaluations.



\begin{figure}[t]

\centering
\includegraphics[width=\linewidth]{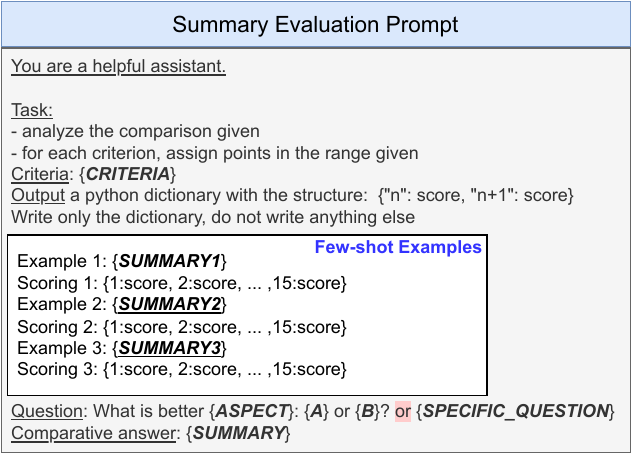}
    \caption{Evaluation prompt for LLMs based on the CompQA evaluation framework criteria. For few-shot, 2 human answers are added to the placeholder.}
    \label{fig:eval_prompt}
    \vspace{-0.2cm}
\end{figure}

\subsection{Experimental Setup}
\paragraph{Participated CQA Agents \& Datasets.}

Here are short descriptions of systems used for competition in the CompQA Benchmark. We use the standard configuration and parameters for generation: 

\begin{itemize}[itemsep=0pt, topsep=0pt]
    \item \textbf{ChatGPT} (gpt-3.5-turbo), details are described by \citet{DBLP:conf/nips/BrownMRSKDNSSAA20};
    \item \textbf{GPT-4o}, more details in \citet{DBLP:journals/corr/abs-2303-08774};
        \item \textbf{meta-llama/Meta-Llama-3-8B}, more details in  \citet{DBLP:journals/corr/abs-2407-21783};
    \item \textbf{meta-llama/Meta-Llama-3-70B}, more details in \citet{DBLP:journals/corr/abs-2407-21783};
    \item \textbf{mistralai/Mixtral-8x7B-Instruct-v0.1} is a pretrained generative Sparse Mixture of Experts \cite{DBLP:journals/corr/abs-2401-04088};
    \item \textbf{Perplexity AI}\footnote{\url{https://www.perplexity.ai}} --- AI-powered research and conversational search engine that answers queries using natural language predictive text, it uses sources from the web and cites links within the text response;
    \item \textbf{CAM 2.0} from \cite{shallouf-etal-2024-cam} based on the \textit{``lmsys/vicuna-7b-v1''} generations on the  Comparative Question Answering pairs on the same pairs;
    \item \textbf{Yahoo!Answers dataset} from \citet{chekalina-etal-2021-better} comprises questions and answers (written by humans) for 28 object pairs that are also presented in our dataset.
    \item \textbf{Human dataset} --- we asked four experts in computational linguistics with at least a bachelor's degree to improve the datasets generated automatically by GPT-4 using Scenario 4. First, they worked on the same 10 answers and discussed the reasoning behind their choices to align their understanding of how the annotation is to be performed. This resulted in 40 manually edited summaries. After that, each expert went on to improve 10 more texts on their own. In total, 80 answers were created. We also found the SBERT metric \cite{reimers-gurevych-2019-sentence} between human and GPT-generated answers $\approx0.92$, and ROUGE \citet{lin-2004-rouge} to be $\approx0.6$. This indicates that GPT-written answers are highly similar to manually created ones.
 
\end{itemize}

More information on the datasets from previous work can be found in Appendix \ref{appendix:data}. For each LLM, four summaries are generated by using different prompt templates. We also return three versions of the answer for the GPT family of models (choices, $n=3$), which are all counted as separate summaries and are all evaluated separately. Overall, we generate \textbf{2,000} (4 models with 4 scenarios and 50 examples each and 2 models with 4 scenarios and 150 examples each) comparative answers.


\paragraph{Obtaining Automatic Assessments using LLMs.}



We automatically evaluated 2,158\footnote{$4scenarios\cdot150_{GPT-3.5} + 4scenarios\cdot150_{GPT-4} + 28_{Yahoo} + 50_{CAM} + 80_{Human} + 4scenarios\cdot50\cdot4_{LLaMA3-8b, LLaMA3-70b, Mixtral, Perplexity} = 2158$} answers with different LLM models (GPT-4, GPT-3.5, Perplexity, Mixtral, LLaMA-3 8b and 70b) based on the CompQA evaluation framework criteria in 2-shot regimes, providing an example of a good and a bad answer, according to human evaluation.  This helps to produce more coherent automatic assessments with the human assessments.

\begin{table*}[ht!]
\centering
\resizebox{0.95\linewidth}{!}{
\begin{tabular}{llllllll}
\toprule
            & \textbf{GPT-4}                                                                                                                  & \textbf{GPT-3.5}                                                                                                & \textbf{Perplexity}                                                                                                            & \textbf{Mixtral}                                                                                                                            & \textbf{LLaMA-3 8b}                                                                                              & \textbf{LLaMA-3 70b}                                                                                   &   \textbf{Human*}                         \\
\midrule
GPT-4        & \textbf{17.77$_{\pm1.06}$}                                                                                                                & \textbf{17.54}$_{\pm2.14}$                                                                                             & \textbf{18.69}$_{\pm0.69}$                                                                                                                & \textbf{18.45$_{\pm2.64}$}                                                                                                                              & \textbf{16.64}$_{\pm1.86}$                                                                                                 & \textbf{18.40}$_{\pm0.70}$    & \underline{15.96}$_{\pm1.35}$                                                                                                            \\
GPT-3.5        & 16.66$_{\pm1.86}$                                                                                                                & 16.05$_{\pm2.62}$                                                                                             & 16.12$_{\pm2.64}$                                                                                                                & 18.05$_{\pm1.33}$                                                                                                                              & 15.08$_{\pm3.18}$                                                                                                 & 16.25$_{\pm2.32}$                                                                  & 14.58$_{\pm2.34}$                                               \\
Perplexity  & \underline{17.42$_{\pm1.34}$}                                                                                                                & 16.52$_{\pm3.00}$                                                                                               & \underline{18.47}$_{\pm1.11}$                                                                                                                 & 18.34$_{\pm0.87}$                                                                                                                              & \underline{16.31}$_{\pm2.06}$                                                                                                 & 17.87$_{\pm1.15}$   & 15.00$_{\pm1.77}$                                                                                                             \\
Mixtral     & 17.20$_{\pm1.31}$                                                                                                                  & \underline{17.05}$_{\pm2.43}$                                                                                              & 17.92$_{\pm1.41}$                                                                                                                & \underline{18.36}$_{\pm0.94}$                                                                                                                             & 16.29$_{\pm2.40}$                                                                                                & 17.45$_{\pm1.67}$                                                                        & 13.75$_{\pm2.63}$                                         \\
LLaMA-3 8b  & 16.58$_{\pm1.80}$                                                                                                                 & 15.95$_{\pm2.60}$                                                                                              & 17.19$_{\pm1.81}$                                                                                                                 & 18.08$_{\pm1.24}$                                                                                                                              & 15.56$_{\pm2.72}$                                                                                                & 16.15$_{\pm1.90}$                                                              & 14.77$_{\pm2.16}$                                                  \\
LLaMA-3 70b & 17.12$_{\pm1.37}$                                                                                                                 & 16.22$_{\pm2.76}$                                                                                               & 17.93$_{\pm1.39}$                                                                                                                 & 18.19$_{\pm1.068}$                                                                                                                             & 15.70$_{\pm2.95}$                                                                                                 & 17.27$_{\pm1.45}$                                                                 & 15.37$_{\pm1.72}$                                               \\
CAM \cite{shallouf-etal-2024-cam}         & 9.52$_{\pm3.808}$                                                                                                                  & 8.74$_{\pm4.56}$                                                                                                & 6.46$_{\pm2.89}$                                                                                                                  & 9.66$_{\pm4.43}$                                                                                                                               & 6.54$_{\pm4.21}$                                                                                                  & 6.40$_{\pm2.29}$                                                                         & 8.04$_{\pm2.78}$                                          \\
Yahoo \cite{chekalina-etal-2021-better}       & 9.32$_{\pm3.95}$                                                                                                                 & 5.88$_{\pm4.86}$                                                                                              & 5.96$_{\pm3.68}$                                                                                                                 & 13.32$_{\pm4.97}$                                                                                                                              & 5.21$_{\pm5.91}$                                                                                                  & 5.96$_{\pm3.64}$                                                               & 6.00$_{\pm1.00}$                                                   \\
Human       & 17.09$_{\pm1.72}$                                                                                                               & 15.29$_{\pm2.68}$                                                                                             & 18.33$_{\pm1.03}$                                                                                                                & 17.65$_{\pm1.57}$                                                                                                                              & 14.29$_{\pm3.23}$                                                                                                & \underline{17.96}$_{\pm1.56}$                                                                   & \textbf{16.69}$_{\pm1.76}$                                            \\
\bottomrule
\end{tabular}}
    \caption{Average scores for all participating models for LLM  and human evaluations. Rows are the agents and datasets. Columns represent the evaluation models. Human* evaluations were conducted on random subset.}
    \label{tab:best_results}
\end{table*}

\paragraph{Obtaining Assessments from Human Annotators.}

To verify the quality of automatic annotation, we randomly selected and manually evaluated a total of \textit{367} answers. \textit{123} of them are done with an overlap of two people to measure the annotation agreement. Krippendorf's alpha is equal to \textbf{0.75} for the final score and \textbf{0.71} for all scores on average. 

\subsection{Comparison of LLM and Human Assessments}

In order to understand whether the human annotation can be replaced with an automatic one, we compare the agreement and annotation scores between different LLMs and human answers. Table \ref{tab:corr} demonstrates both the agreement score (Krippendorff's alpha) and Spearman correlation scores, following  \citet{DBLP:journals/corr/abs-2406-18403}. The results give a positive answer to the \textbf{RQ-2}: 
``LLMs can reliably evaluate comparative summaries with human expert-level quality.''
\begin{table}[t]
    \centering
    \vspace{-0.5cm}\resizebox{0.95\linewidth}{!}{
    \begin{tabular}{lcc}
    \toprule
     \textbf{Model}    & \textbf{$\alpha$} & \textbf{Spearman's} \\
      \midrule
      GPT-4, separately   & \textbf{0.71} & 0.69, $p<0.001$\\
      GPT-4, final score  & \textbf{0.58} & 0.55, $p<0.001$ \\
      \midrule
      GPT-3.5, separately   & \underline{0.63} & 0.63, $p<0.001$\\
      GPT-3.5, final score  & 0.31 & 0.40, $p<0.002$ \\
      \midrule
      Perplexity, separately   & \textbf{0.71} & \underline{0.72}, $p<0.001$\\
      Perplexity, final score  & 0.39 & 0.60, $p<0.001$ \\
      \midrule
      Mixtral, separately   & 0.55 & 0.69, $p<0.001$\\
      Mixtral, final score  & 0.22 & \underline{0.62}, $p<0.001$ \\
      \midrule
      LLaMA-3 8b, separately   & 0.48 & 0.49, $p<0.001$\\
      LLaMA-3 8b, final score  & 0.41 & 0.46, $p<0.001$ \\
      \midrule
      LLaMA-3 70b, separately   & 0.61 & \textbf{0.76}, $p<0.001$\\
      LLaMA-3 70b, final score  & \underline{0.47} & \textbf{0.72}, $p<0.001$ \\
      
    \bottomrule
    \end{tabular}}
    \caption{Agreement (Krippendorf's $\alpha$) and correlation scores between human and LLM evaluations. Separate scores are calculated for all scores concatenated for all answers (human annotation against model annotation), the total scores represent the sums of 15 criteria (denoted in Figure 3) for each answer.}
    \label{tab:corr}
\end{table}

\begin{figure}[!ht]
\centering
\vspace{-0.5cm}
\includegraphics[width=\linewidth]{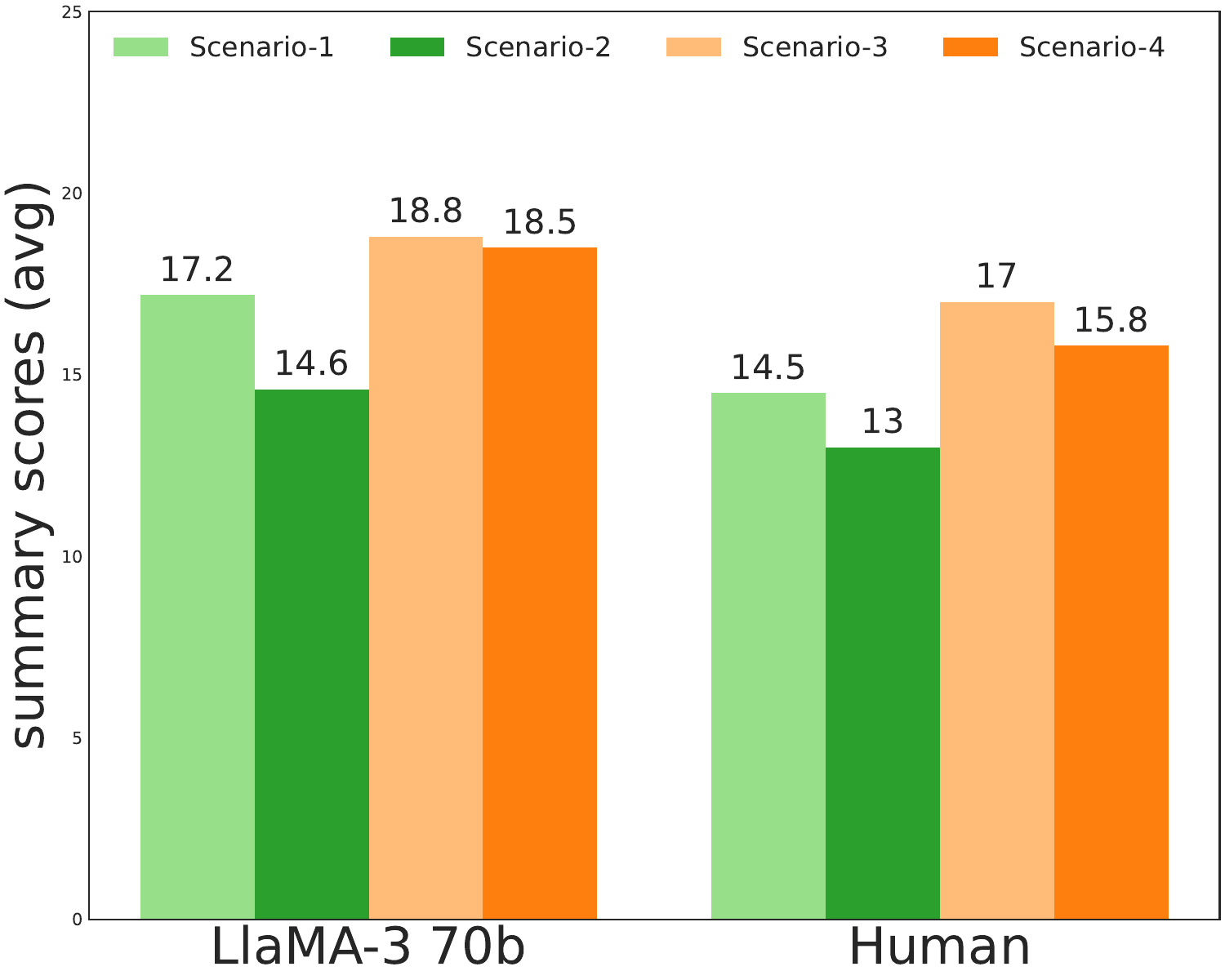}
    \caption{Average scores for each scenario for LLaMA-3 70b and manual human assessments).}
    \label{fig:scenario_comparison}
    \vspace{-0.4cm}
\end{figure}

From the results in Table \ref{tab:corr}, we can see that both agreement and correlation scores are much higher for the separate scores than for the summary scores. Moreover, we can also conclude that LLaMA-3 70b is better according to Spearman's correlation, while GPT-4 and Perplexity show higher agreement according to Krippendorf's $\alpha$.











\begin{figure*}[ht!]
\vspace{-0.5cm}
    \centering
    \includegraphics[ width=0.8\linewidth]{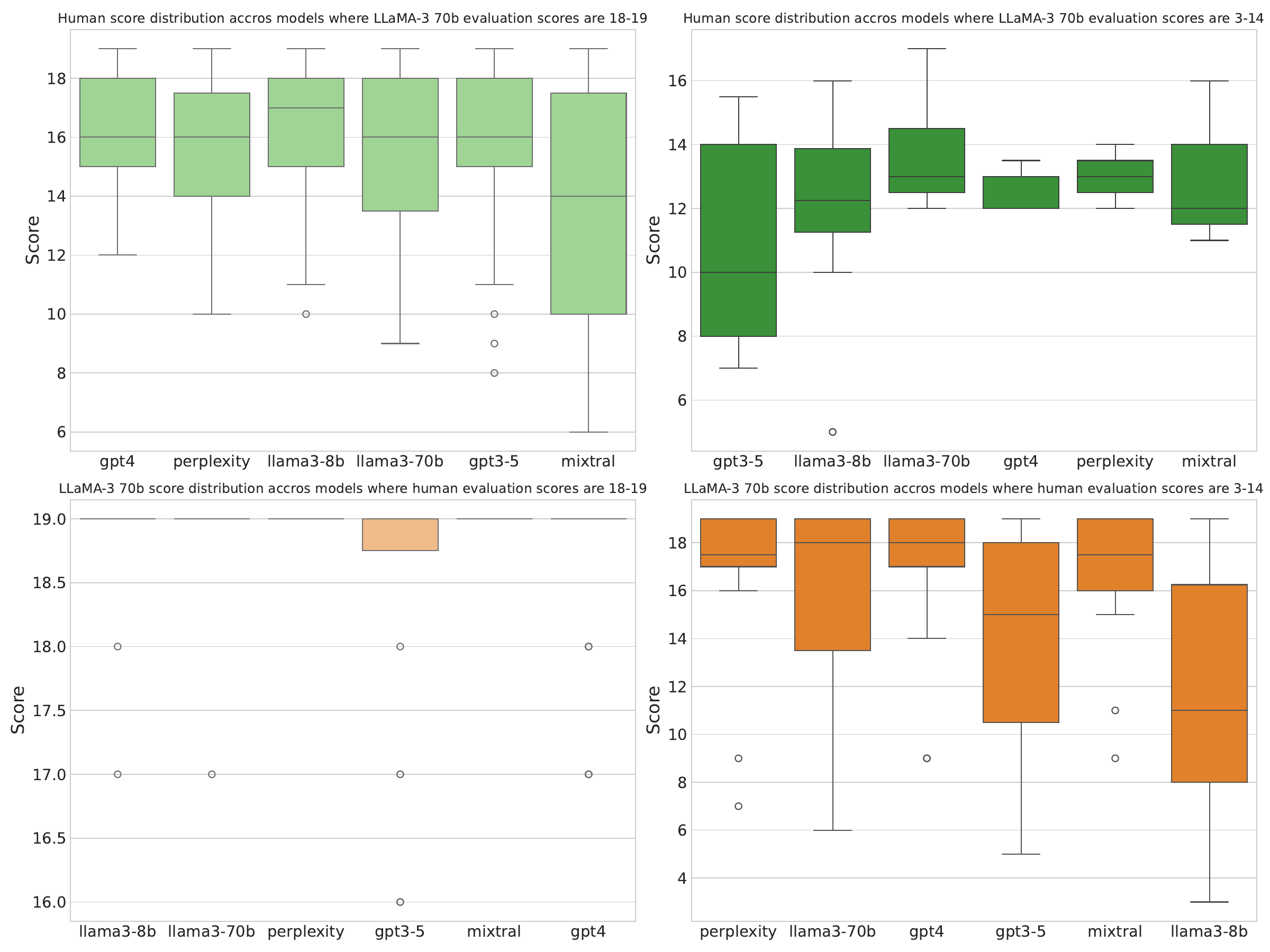}
    \caption{Distribution scores across models for the highest and lowest scores assigned by LLaMA-3 70b and Human evaluations.}
    \label{fig:analysis}
\end{figure*}

\section{CQA Systems Performance}

In this section, we present the results of existing CQA agents on our benchmark, and answer the \textbf{RQ3}: ``How do different LLMs fare against each other in generating high-quality summaries?''
Table~\ref{tab:best_results} shows the best results obtained by each model (row) on our benchmark with different evaluators (column).  For human evaluations, scores are reported for a randomly sampled dataset of 367 summaries. 
The results of all LLM evaluators unanimously rate the GPT-4 answers as the first. The second place is shared between Mixtral and Perplexity.
Regarding human evaluations, it is evident that the highest quality responses are those generated by humans (\textbf{16.69} on average), while the best LLM answers are still generated by GPT-4. Interestingly, the answers ranked as the lowest belong to Yahoo and CAM datasets.
To see the detailed results, refer to Table~\ref{tab:results_all} in Appendix \ref{appendix:complete_results}.

\begin{figure}[ht!]
\centering
\vspace{-0.5cm}
\includegraphics[width=\linewidth]{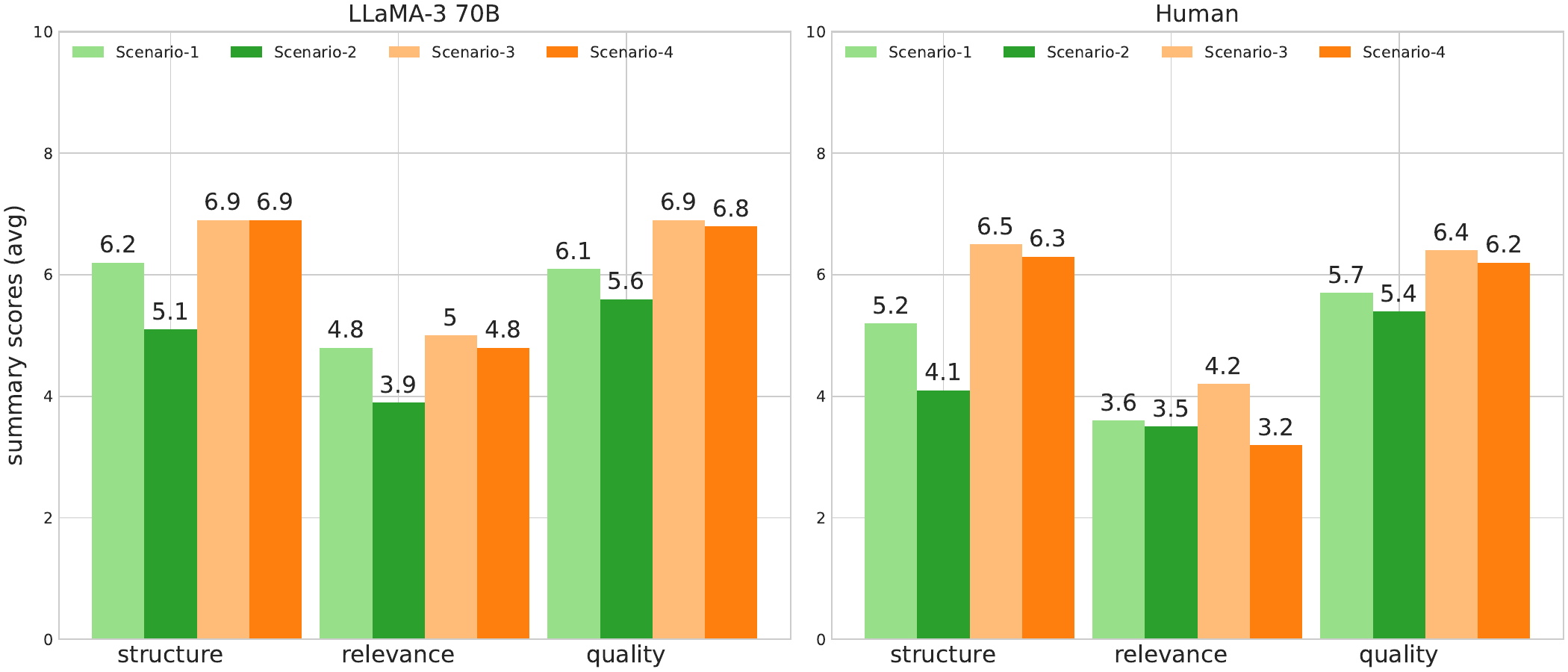}
    \caption{Average scores for each scenario, distributed into three categories of evaluation framework:  \textbf{structure}--(7 points),  \textbf{quality}--(5 points), and \textbf{relevance}--(7 points).}
    \label{fig:criteria_comparison}
    \vspace{-0.5cm}
\end{figure}

\paragraph{Comparison between scenarios.} 
Figure~\ref{fig:scenario_comparison} shows average scores for all models. Scenarios 3 and 4 have the highest scoring for both human and LLaMA-3 70b evaluation. The summaries generated with the first and 2nd prompt scenarios are consistently ranked lowest. This highlights the fact that the evaluation framework can help to differentiate between summaries of varied structures. The fourth scenario uses arguments from CAM, and they have scored lower than their similar counterpart without these arguments i.e. the third scenario, which is more distinct on the human evaluation. This discrepancy raises concerns about the quality of CAM arguments, despite the prompt explicitly requiring the selection of only relevant arguments. Thus, we further compare these models for all criteria of our evaluation framework, by grouping them into \textit{structure}, \textit{quality}, and \textit{relevance} in Firgure~\ref{fig:criteria_comparison}. The comparison shows that the summaries generated with the third and fourth scenarios scored higher for all categories of our framework. The second scenario suffered noticeably for \textit{structure}, this difference is less noticeable between the third and fourth scenarios.


\paragraph{Comparison between LLaMA-3 70b and human:}
Human evaluations are consistently lower than LLaMA-3 70b. 
Table~\ref{tab:aspect_comp_results} shows that LLaMA-3 70b performs similarly for the questions with and without aspect, while the human scores are lower when the aspect is presented, which implies that LLaMA-3 70b might not  give the importance to the required aspect when evaluating the summary.

Additional analysis is also present in Figure \ref{fig:analysis}: we aim to check, whether the scores assigned by LLaMA-3 70b to the answers evaluated by humans for the extreme cases are coherent. 
Human evaluation shows average results (with means at 14-17) for the cases evaluated as good ones by LLaMA-3 70b and also low results for the answers assigned with low scores by the LLM. It means that the model slightly tends to overestimate the performance of the LLMs in comparison to the human evaluation.
From the second row, we can see that can see that LLaMA-3 70b assigns high scores for the answers which are also ranked high by humans (lower left subfigure) and both high and low scores for the answers ranked as ``bad'' (3-14) by humans (lower right subfigure). 

\begin{table}[t]
    \centering

    \begin{tabular}{lrr}
\toprule
\multicolumn{1}{c}{\multirow{1}{*}{\textbf{Aspect}}} &  \multirow{1}{*}{\textbf{LLaMA-3 70b}}&   \multirow{1}{*}{\textbf{Human}}     \\
\midrule
yes  & 16.43$_{\pm3.52}$ & 13.43$_{\pm2.98}$ \\
no & 16.44$_{\pm3.52}$ & 15.01$_{\pm2.69}$ \\
\bottomrule
    \end{tabular}
    \caption{Average scores of human and LLaMA-3 70b, for the subset of answers \textbf{with} and \textbf{without} aspect.}
    \vspace{-0.5cm}\label{tab:aspect_comp_results}
\end{table}

\paragraph{Possible Biases Inherent to Certain LLMs.}
Another possible direction that should be worked on to improve upon this research paper is finding any biases that could exist within some LLMs. 

An example of this that we encountered while generating summaries was that GPT-4, when asked to compare Google and Yahoo search engines and given arguments favoring the latter, created comparisons strongly for Google (see example in Figures \ref{fig: google} and \ref{fig:yahoo}). Moreover, the model rewrote the arguments, hallucinating that they were all originally favoring Google. We argue that such cases of bias where the final answer is not derived from the listed arguments could be because of the model inner bias towards a more popular object or subject.

Needless to say, this kind of favoritism embedded in an LLM may heavily impact its objectivity, and further tests need to be conducted to determine the cause and possible solutions to this issue. This is an exception, however, as no other object pair created such a situation; this means that, hopefully, this clear bias is something exceedingly rare.

\section{Related Work}
In this section, we briefly introduce each subtask for Comparative Question Answering and also discuss the existing LLM evaluation benchmarks.


\subsection{Comparative Question Answering}

Here, we introduce each subtask and list several papers that addressed the topic.

\textbf{Comparative Question Identification} aims at classifying questions into two types: comparative and non-comparative. This classification task is solved with both Encoder and Decoder Transformer models \cite{bondarenko2020comparative,DBLP:conf/wsdm/BondarenkoADHBH22,shallouf-etal-2024-cam}.
\textbf{Object and Aspect Identification} is a sequence labelling task, aims at finding objects and aspect of comparison in the question. There exist varios datasets and approaches to solve the task, mostly, with Trasnformer models \cite{chekalina-etal-2021-better,beloucif-etal-2022-elvis,DBLP:conf/wsdm/BondarenkoADHBH22,shallouf-etal-2024-cam}.
\textbf{Stance Classification} is another classification task, that identifies the stance of comparative sentences.  \citet{panchenko-etal-2019-categorizing}, \citet{DBLP:conf/wsdm/BondarenkoADHBH22}, and \citet{DBLP:journals/corr/abs-2310-08523} solve the task using standard ML classifier, Encoder-based Transformer, and GPT-4 respectively.
\textbf{Summary Generation} is only partially tackled by \citet{chekalina-etal-2021-better} and \citet{shallouf-etal-2024-cam}. The closest work on multi-document summarization of differing opinions is by \citet{iso-etal-2022-comparative}, which focuses on aggregating diverse opinions and synthesizing them into a coherent summary.

\subsection{LLM Evaluation Benchmarks}
 


Apart from the well-known benchmarks like SuperGLUE \cite{sarlin2020supergluelearningfeaturematching}, MTEB \cite{muennighoff-etal-2023-mteb}, and SQuAD \cite{rajpurkar2016squad100000questionsmachine}, several more challenging benchmarks have gained prominence: MMLU (Massive Multitask Language Understanding) \cite{DBLP:conf/iclr/HendrycksBBZMSS21},  TruthfulQA \cite{lin-etal-2022-truthfulqa}, BIG-Bench Hard \cite{suzgun-etal-2023-challenging}. The LLM evaluation framework proposed by \cite{chiang-lee-2023-large} involves presenting a Large Language Model with task instructions, a sample to be evaluated and a question. The researchers use LLM evaluation to score parts of both generated and human-written stories. 
To facilitate comprehensive evaluations, several initiatives aggregate multiple benchmarks:
Hugging Face's Big Benchmarks Collection (a centralized platform for various leaderboards), and 
LMSys Chatbot Arena \cite{chiang2024chatbot,zheng2023judgingllmasajudgemtbenchchatbot}: it also utilizes user ratings and GPT-4 evaluations to assess chatbot performance.

\section{Conclusion}


We have proposed a comprehensive CompQA evaluation framework based on 15 various criteria to be used in scoring comparative answers, demonstrating that GPT-4 produces the best answers to comparative questions. We have manually evaluated the answers from a range of models and datasets using our framework and compared human scores with LLM evaluations, showing that LLM results have a strong correlation with human expert-level evaluations. 
As part of our future work, we plan to enlarge the comparative question dataset and train smaller language models for this task. 

\section{Limitations}
The main limitations of the paper are as follows:

\begin{itemize}

\item Regarding the human evaluation dataset, we acknowledge that our sample size is quite small and needs to be extended to make comparisons of better quality. 

\item Another limitation is the number of models tested. Due to time constraints, we could not add more LLMs to our experiments and leave it for future work. 

\item The arguments used in the strategies are solely derived from the CAM framework. This reliance on a single source may have constrained the retrieval performance, particularly in Scenario 2 and 4, leading to suboptimal results. To address this, future work could involve annotators in crafting or refining arguments, which may enhance the robustness and effectiveness of the strategies.

\end{itemize}

\section{Ethical Considerations}

In our benchmark we test multiple LLMs, including ChatGPT and GPT-4. One key concern is the handling of user data by proprietary models like OpenAI's GPT-4, which are developed and maintained by private companies. These companies often retain the right to use input data for improving their models, as stated in their terms of service and privacy policies. As a result, personal or sensitive information provided by users during interactions with these models could be logged, stored, or analyzed for commercial purposes. While companies may anonymize data, the potential use of personal information in ways that users may not fully understand or consent to raises significant privacy concerns.

\bibliography{custom2.bib}

\appendix

\onecolumn

\section{Full Evaluation Prompt}\label{appendix:eval}

\begin{figure}[th!]
    \centering
    \begin{lstlisting}
    You are a helpful assistant.
    Task:
    - analyze the comparison given
    - for each criterion, assign points in the range given
    Criteria:
    1. a short introduction is present: 0-1
       - the introduction is missing or is too long - 0 points
       - the introduction is short and concise - 1 point
    2. there are defined aspects used for comparison: 0-1
       - the comparison is arbitrary with no specific aspects - 0 points
       - the summary uses specific aspects to compare objects - 1 point
    3. the introduction mentions the most important comparison aspects: 0-1
       - no aspects are mentioned or no introduction - 0 points
       - several most important aspects are mentioned in the introduction - 1 point
    4. the main body of the comparison has good structure: 0-1
       - some aspects mix with others, the structure is harder to follow - 0 point
       - the aspects are logically divided into separate aspects - 1 point
    5. the main body of the comparison has defined aspect names: 0-1
       - no aspect names are given, comparison is inconcrete - 0 points
       - main body has distinct aspect names - 1 point
    6. the main body of the comparison has defined aspect descriptions: 0-1
       - no aspect descriptions are given, comparison is inconcrete - 0 points
       - main body has distinct aspect descriptions - 1 point
    7. the final choice is given explicitly: 0-1
       - no explicit choice made or lengthy justification present - 0 points
       - short and explicit choice made - 1 point
    8. the comparison aspects in the main body of the comparison are sorted by general applicability: 0-1
       - statements are not sorted at all - 0 points
       - statements are sorted by general/important statements first, specific statements closer to the end - 1 point
    9. each argument is relevant to the subject of comparison: 0-2
       - most arguments are irrelevant - 0 points
       - most arguments are relevant - 1 point
       - all arguments are relevant - 2 points
    10. each argument compares both objects: 0-2
       - some arguments do not compare the objects - 0 points
       - some arguments give information only about one object - 1 point
       - all arguments compare both objects - 2 points
    11. there are no hallucinations or statements contradicting common knowledge: 0-2
       - many hallucinations, serious factual inaccuracy - 0 points
       - some hallucinations, but mostly correct - 1 point
       - no hallucinations, factually correct - 2 points
    12. the comparison has proper language and is easy to follow: 0-2
       - hard to read, profanity present or illogical - 0 points
       - some grammar issues, broken logic - 1 point
       - no grammar issues, good structure and logic - 2 points
    13. there are no repetitive statements or statements too similar to each other: 0-1
       - some statements repeat others' meaning very closely - 0 points
       - all statements are unique and do not repeat - 1 point
    14. the final answer is concluded from the statements in the main body and takes the main aspect (if there is one) into consideration (if all statements favor object 1, then the answer is object 1, if both objects are equally good or equally bad, then none of the objects is preferred and the answer is inconclusive): 0-1
       - the final answer is not concluded from the arguments or main aspect (if there is one) or no answer is given - 0 points
       - the final answer is concluded from the majority of arguments and main aspect (if there is one) - 1 point
    15. the summary itself is not too short and not too long: 0-1
       - the summary is too short (less than 12 sentences) or too long (more than 20 sentences) - 0 points
       - the summary is reasonably long (from 12 to 20 sentences) - 1 point
    \end{lstlisting}
    \caption{Part 1: Full Evaluation Prompt used for directing an LLM to analyze comparative summaries in line with our 15 criteria. This example also includes a sample summary evaluation for few-shot prompting.}
    \label{fig:full_eval_prompt}
\end{figure}

\begin{figure}[ht!]
    \centering
    \begin{lstlisting}
        Output a python dictionary with the structure: {"n": score, "n+1": score}
    Write only the dictionary, do not write anything else
    Example 1:
    Summary:
    The Nintendo DS and PlayStation 3 are two iconic gaming consoles that revolutionized the gaming industry. While the DS focused on handheld gaming, the PS3 offered a high-definition gaming experience on a home console. Both devices have their unique features and strengths, making it challenging to determine the better option.
    Main Aspects of Comparison:
    - Graphics: The PS3 boasts superior graphics capabilities with its high-definition output, providing a more immersive gaming experience compared to the DS's smaller screen.
    - Game Library: The PS3 has a vast library of games, including popular titles like Uncharted" and "The Last of Us", while the DS offers a wide range of unique and innovative games such as "Mario Kart DS" and "The Legend of Zelda: Phantom Hourglass."
    - Online Capabilities: The PS3's online platform, PlayStation Network, offers robust multiplayer options and digital downloads, whereas the DS's online features are more limited.
    - Portability: The DS's handheld design allows for gaming on the go, making it a convenient option for travel or commutes, while the PS3 is limited to home use.
    - Controller Design: The DS features a touchscreen and stylus for innovative gameplay, while the PS3's DualShock controller provides a more traditional gaming experience.
    Best Option:
    PS3
    Scoring 1:
    {1: 1,	2: 1, 3: 1, 4: 1, 5: 1, 6: 1, 7: 1, 8: 1, 9: 2, 10: 2, 11: 2, 12: 2, 13: 1, 14: 1, 15: 1}
    Question: What is better{aspect_text}: {object1} or {object2}?
    Comparative answer:
    {comparison}
    \end{lstlisting}
    \caption{Part 2: Full Evaluation Prompt used for directing an LLM to analyze comparative summaries in line with our 15 criteria. This example also includes a sample summary evaluation for few-shot prompting.}
    \label{fig:full_eval_prompt2}
\end{figure}

\clearpage
\section{Example Summary Evaluation} \label{appendeix:example_summary_eval}
Figure~\ref{fig:ex_summary} shows the summary produced by \textbf{Mixtral} model to compare \textbf{dell} and \textbf{ibm} for their \textbf{power of processors}. The summary was generated using the \textbf{third} prompt scenario. 
\subsection{Summary}

\begin{figure*}[th!]
    \centering
\begin{lstlisting}
Summary:
Dell and IBM are technology giants, offering a diverse range of products and services. Dell, a prominent hardware manufacturer, provides a wide array of customizable PCs, laptops, and servers, often praised for their affordability and ease of use. IBM, on the other hand, has a strong focus on enterprise solutions, offering a comprehensive suite of software, services, and consulting. IBM is known for its robust mainframe computers and cognitive computing solutions, leveraging AI and analytics. Both companies have a global presence and cater to a variety of customers, from individuals to large enterprises. This comparison will focus on their key differences to determine which offers the better overall package. 

Comparison:

- Product Range:
   - Dell: Offers a broad range of customizable PCs, laptops, and servers for consumers and businesses. Known for affordability, ease of use, and flexibility in specifications.
   - IBM: Focuses on enterprise solutions, including mainframe computers, software, and cognitive computing. Provides robust hardware and advanced AI, analytics, and cloud services.

- Target Market:
   - Dell: Caters to a wide range of consumers, small businesses, and enterprises. Offers products suitable for various budgets and needs.
   - IBM: Primarily targets large enterprises and organizations with its comprehensive suite of advanced solutions.

- Strengths:
   - Dell:
     - Affordability: Known for providing quality products at competitive prices.
     - Customizability: Allows users to customize hardware specifications to meet specific needs.
     - Ease of Use: Dell products are often user-friendly and intuitive.

   - IBM:
     - Enterprise Solutions: Offers a one-stop shop for enterprises with advanced hardware, software, and consulting services.
     - Innovation: Invests heavily in R&D, leading to innovative solutions like Watson cognitive computing.
     - Mainframe Expertise: Renowned for robust and reliable mainframe computers.

Best Option: 
Tie
\end{lstlisting}
\caption{Example Summary.  generator model:\textbf{Mixtral} --- prompt scenario: \textbf{3} --- object: \textbf{dell} and \textbf{ibm} --- aspect:\textbf{power of processors}. }
\label{fig:ex_summary}
\end{figure*}

\begin{figure*}[th!]
 \centering
\begin{lstlisting}
**Summary:** 
When comparing Google Search and Yahoo, Google's dominance in the search engine market is undeniable due to its superior user experience, relevance of search results, integration with other services, and overall brand reputation.  
**Main Aspects of Comparison:** 
- User Interface: Google features a clean and minimalist design, while Yahoo search interface is more cluttered and more overwhelming [1,4]. 
- Search Quality: Google's search algorithm is renowned for its accuracy and relevance in delivering search results, outperforming Yahoo. 
- Search speed: Google search is known for its fast search results than Yahoo - Market Dominance: Google dominates the search engine market, giving users access to a vast database of information, in comparision to the Yahoo search, struggling to compete with Google. 
- User experience: Both Google and Yahoo Search are praised for their browsing experience [1]. 
- Features & Services: Google offers a wider range of integrated services and functionalities than Yahoo. [4] Both Google and Yahoo search may offer unique features or tools that enhance the search experience.
 - Mobile compatibility & Cross-platform integration: Google Search is more optimized for mobile devices, offering a better integration across various platforms than Yahoo Search.  
**Best Option:** Google
\end{lstlisting}
    \caption{Example of a human summary, objects: \textbf{Google} and \textbf{Yahoo}.}
    \label{fig: google}
\end{figure*}

\begin{figure*}[th!]
\centering
\begin{lstlisting}
Summary: 
Google Search and Yahoo Search offer distinct user experiences, each with its strengths and weaknesses. While Google dominates the market, Yahoo Search presents competitive features and a pleasant interface, particularly highlighted in its integration with services like Flickr.  
Main Aspects of Comparison: 
- User experience: Yahoo search is considered to provide a nicer experience compared to Google search [1] 
- Past performance: Yahoo search was deemed better than Google search in the first half of 2008 [2] 
- Search quality: Yahoo search delivers more relevant results than Google search, highlighting its superiority [3], although one may argue that Google's search algorithms are often considered more sophisticated and effective [generated]. 
- Unique services: Yahoo's Flickr photo service surpasses Google's offerings, suggesting a wider range of services [4] 
- Visual appeal: Yahoo search may have a more visually appealing interface compared to Google search [generated] 
- Additional features: Yahoo search may offer unique features or tools that enhance the search experience [generated] 
- Search speed: Google search is known for its fast search results, which may be a crucial factor for some users [generated] 
- Customization Options: Yahoo Search provides extensive customization options, allowing users to tailor their search preferences. [generated] 
- Accessibility: Google Search is more widely accessible across various devices and platforms [generated].    
Best option: Yahoo
\end{lstlisting}
    \caption{Example of another human summary, objects: \textbf{Google} and \textbf{Yahoo}. }
    \label{fig:yahoo}
\end{figure*}

\clearpage
\subsection{Evaluation}

Table~\ref{tab:ex_summary_eval} shows the evaluation of the example summary across all 15 criteria, done by a human expert and LLaMA-3 70b. For criterion C1, the introduction of the summary is pretty long and marked \textit{0} by the human expert, but \textit{1} by the LLaMA-3 70b, this is also a tricky decision in the absence of description in terms of word or sentence count.
For the sorting of arguments (C8), LLaMA-3 70b assigned \textit{0}, while human annotation sees the aspects as sorted by general applicability. The last difference is the C9 criterion: the human annotation is stricter and assigns \textit{0}, while LLaMA-3 70b assumes that all arguments are relevant.



\begin{table*}[htbp]
  \centering
  \small
   
    \begin{tabular}{p{5mm}|p{10cm}|m{16mm}m{8mm}}
    \toprule
    \multirow{2}[0]{*}{No.} & \multirow{2}[0]{*}{Criteria description} & \multicolumn{2}{c}{Evaluation}\\ 
     & &  LLaMA-3 70b & Human \\
     \midrule
    C1     & a short introduction is present: \textbf{0-1} & 1          & \underline{0} \\
    C2     & there are defined aspects used for comparison in the whole comparison: \textbf{0-1}      & 1     & 1 \\
    C3     & the introduction mentions the most important comparison aspects: \textbf{0-1} & 1          & \underline{0} \\
    C4     & the main body of comparison has good structure: 0-1 & 1         & 1 \\
    C5     & the main body of the comparison has defined aspect names: \textbf{0-1} & 1         & 1 \\
    C6     & the main body of the comparison has defined aspect descriptions: \textbf{0-1}     & 1     & 1 \\
    C7     & the final choice is given explicitly: \textbf{0-1}     & \underline{0}     & \underline{0} \\
    C8     & the comparison aspects in the main body of the comparison are sorted by general applicability: \textbf{0-1}      & \underline{0}     & 1 \\
    C9     & each argument is relevant to the aspect of comparison \newline{}(if any, otherwise is general and is not biased towards any aspect): \textbf{0-2}     & 2     & \underline{0} \\
    C10    & each argument compares both objects: \textbf{0-2}      & 2     & 2 \\
    C11    &  here are no hallucinations or statements contradicting common knowledge: \textbf{0-2}      & 2     & 2 \\
    C12    & the comparison has proper language and is easy to follow: \textbf{0-2}     & 2     & 2 \\
    C13    & here are no repetitive statements or statements too similar to each other: \textbf{0-1}      & 1     & 1 \\
    C14    & the final answer is concluded from the statements in the main body and takes the main aspect (if there is one) into consideration: \textbf{0-1}       & \underline{0}     & \underline{0} \\
    C15    & the summary itself is not too short and not too long: \textbf{0-1}      & 1     & 1 \\
    \bottomrule

    \end{tabular}
    \caption{Evaluation of summary from Figure~\ref{fig:ex_summary}. \textbf{underline} highlight the disagreement for the corresponding criteria.}
  \label{tab:ex_summary_eval}
\end{table*}%

 \clearpage

\section{Datasets Details}\label{appendix:data}

In addition to the answers created with 6 models and four prompt scenarios, we also assess previously available datasets. 

\begin{itemize}
    \item \textbf{Touché Dataset} \cite{bondarenko:2022e} is a dataset created for the Touché Shared task on comparative questions. Given a comparison search topic with two comparison objects and a collection of text passages, the task was to retrieve relevant argumentative passages for one or both objects, and to detect the passages' stances with respect to the objects. The authors provided 50 search topics that described scenarios of personal decision making. Each of these topics had a title in terms of a comparative question, comparison objects for the stance detection of the retrieved passages, a description specifying the particular search scenario, and a narrative that served as a guideline for the assessors.
    
    \item \textbf{Yahoo!Answer Dataset} \cite{chekalina-etal-2021-better} comprises 28 answers. The authors use information extracted from Yahoo! Answers: they collect a subset of L6–Yahoo! Answers Comprehensive Questions and Answers version 1.0 retrieved from Yahoo! Webscope. 

    \item \textbf{CAM 2.0 dataset} \cite{shallouf-etal-2024-cam} is an automatically created dataset of comparative answers using the \textit{``lmsys/vicuna-7b-v1.5''} model \cite{vicuna} using a two-shot setup. The authors ask the model to write a comparison summary of objects and also provide a list of arguments extracted from CAM 2.0. The task is to summarize only relevant arguments and to put citations of the arguments inside the generated text to prevent hallucination. The number of summaries is 50.
\end{itemize}

Table \ref{tab:examples_data} provide the examples of the samples from the datasets. 

\begin{table}[ht!]
    \centering
    \begin{tabular}{p{0.3\linewidth}p{0.7\linewidth}}
    \toprule
       Dataset name  & Example \\
       \midrule
       Touché dataset & \{"question": "Which is better, a laptop or a desktop?", "pair": ["laptop", "desktop"]\}
\\
       Yahoo!Answer Dataset  & How can you even ask this question yet? Only the Xbox 360 is out at the moment and that hasn't even been tested by gamers enough to see truely how good or rubbish it is. You need to ask the question again when all three systems are out! \\
       CAM 2.0 dataset  & Microsoft and Sony are two major companies in the technology industry, with a significant presence in the gaming market.
Some argue that Microsoft is better and faster than Sony, with updates going smoother and less frequent [2]. Microsoft is also considered to have a better SDK for games [3], and their conference was thought to have better pacing [4]. Additionally, some believe that Microsoft has a superior position over Sony in terms of software tools [9]. However, others argue that Sony is a superior hardware manufacturer, much better than both Nintendo and Microsoft [11]. The PS4 is physically superior to the Xbox One, with better design [12]. Sony is also believed to be working harder for gamers than Microsoft and Nintendo [10].
In terms of gaming, some argue that Microsoft is inferior to Sony [13], and that Sony is superior to Microsoft in every way possible [17]. Sony was also considered to be a bit smarter than Microsoft in terms of their approach to gaming [18].
Ultimately, the preference between Microsoft and Sony comes down to personal opinions and experiences. Some may prefer Microsoft for its software tools and updates, while others may prefer Sony for its hardware design and gaming experience.
Arguments used: 1,2,3,4,5,6,7,8,9,10,11,12,13,14,15,16,17,18,19,20\\
         \bottomrule
    \end{tabular}
    \caption{Samples from the external datasets assessed in the Comparative QA framework.}
    \label{tab:examples_data}
\end{table}

\clearpage
\section{Complete Results}\label{appendix:complete_results}
Table~\ref{tab:results_all} shows the performance of all models, for each scenario.

\begin{table}[ht!]
\centering
\resizebox{\textwidth}{!}{
\begin{tabular}{lcccccccc}
\toprule
\textbf{Model} & \textbf{Scenario} & \textbf{Mixtral} & \textbf{GPT-4}& \textbf{Human} & \textbf{LlaMA-3 8b} & \textbf{LlaMA-3 70b}  & \textbf{Perplexity} & \textbf{GPT-3.5} \\
\midrule
GPT-3.5 & 1 & 18.76$_{\pm0.59}$ & 16.95$_{\pm1.44}$ & 14.93$_{\pm2.84}$ & 17.07$_{\pm2.14}$ & 16.63$_{\pm1.27}$ & 16.84$_{\pm2.14}$ & 17.90$_{\pm1.65}$ \\
GPT-3.5 & 2 & 16.77$_{\pm2.80}$ & 14.10$_{\pm3.15}$ & 11.46$_{\pm 2.93}$ & 11.86$_{\pm4.00}$ & 12.14$_{\pm4.99}$ & 11.07$_{\pm5.79}$ & 13.39$_{\pm4.07}$ \\
GPT-3.5 & \textbf{3} & 18.71$_{\pm0.57}$ & 18.21$_{\pm1.02}$ & 16.92$_{\pm1.90}$ & 16.59$_{\pm3.09}$ & 18.54$_{\pm1.02}$ & 18.44$_{\pm 1.12}$ & 17.74$_{\pm1.56}$ \\
GPT-3.5 & 4 & 18.02$_{\pm1.35}$ & 17.37$_{\pm1.83}$ & 15.00$_{\pm1.70}$ & 14.81$_{\pm3.51}$ & 17.69$_{\pm2.00}$ & 18.12$_{\pm1.53}$ & 15.17$_{\pm3.21}$ \\
\midrule
GPT-4 & 1 & 18.57$_{\pm0.72}$ & 17.59$_{\pm1.00}$ & 15.22$_{\pm1.52}$ & 17.44$_{\pm1.53}$ & 17.90$_{\pm0.86}$ & 18.41$_{\pm0.88}$ & 17.84$_{\pm2.15}$ \\
GPT-4 & 2 & 17.97$_{\pm8.75}$ & 17.22$_{\pm1.33}$ & 14.48$_{\pm1.62}$ & 14.56$_{\pm2.78}$ & 17.75$_{\pm1.55}$ & 18.41$_{\pm1.50}$ & 16.77$_{\pm2.76}$ \\
GPT-4 & \textbf{3} & 18.91$_{\pm0.31}$ & 18.23$_{\pm0.95}$ & 17.24$_{\pm1.33}$ & 17.71$_{\pm1.53}$ & 18.95$_{\pm0.32}$ & 18.94$_{\pm0.31}$ & 18.36$_{\pm1.14}$ \\
GPT-4 & 4 & 18.38$_{\pm0.78}$ & 18.03$_{\pm0.95}$ & 16.89$_{\pm0.94}$ & 16.83$_{\pm1.59}$ & 18.99$_{\pm0.08}$ & 18.99$_{\pm0.08}$ & 17.21$_{\pm2.47}$ \\
\midrule
LLaMA-3 70b & 1 & 18.74$_{\pm0.56}$ & 17.22$_{\pm1.20}$ & 14.04$_{\pm1.59}$ & 17.34$_{\pm2.16}$ & 17.34$_{\pm1.00}$ & 18.06$_{\pm1.00}$ & 17.38$_{\pm2.66}$ \\
LLaMA-3 70b & 2 & 16.44$_{\pm2.62}$ & 15.90$_{\pm1.78}$ & 13.29$_{\pm2.40}$ & 12.70$_{\pm4.11}$ & 13.86$_{\pm4.22}$ & 15.74$_{\pm4.30}$ & 13.69$_{\pm4.11}$ \\
LLaMA-3 70b & \textbf{3} & 18.76$_{\pm0.59}$ & 18.06$_{\pm1.11}$ & 17.69$_{\pm1.58}$ & 16.82$_{\pm2.67}$ & 18.96$_{\pm0.20}$ & 19.00$_{\pm0.00}$ & 17.90$_{\pm1.31}$ \\
LLaMA-3 70b & 4 & 18.80$_{\pm0.49}$ & 17.30$_{\pm1.37}$ & 16.46$_{\pm1.30}$ & 15.96$_{\pm2.86}$ & 18.90$_{\pm0.36}$ & 18.92$_{\pm0.27}$ & 15.94$_{\pm2.95}$ \\
\midrule
LLaMA-3 8b & 1 & 18.70$_{\pm0.71}$ & 16.78$_{\pm1.40}$ & 14.00$_{\pm2.94}$ & 16.31$_{\pm2.36}$ & 16.52$_{\pm1.71}$ & 17.56$_{\pm0.93}$ & 17.67$_{\pm2.15}$ \\
LLaMA-3 8b & 2 & 16.12$_{\pm3.15}$ & 14.10$_{\pm3.36}$ & 11.94$_{\pm2.65}$ & 11.51$_{\pm4.90}$ & 11.04$_{\pm4.40}$ & 13.52$_{\pm5.41}$ & 11.76$_{\pm4.83}$ \\
LLaMA-3 8b  & \textbf{3} & 18.92$_{\pm0.27}$ & 18.26$_{\pm0.88}$ & 16.50$_{\pm1.96}$ & 18.60$_{\pm0.64}$ & 19.00$_{\pm0.00}$ & 19.00$_{\pm0.000}$ & 18.58$_{\pm0.70}$ \\
LLaMA-3 8b & 4 & 18.59$_{\pm0.81}$ & 17.18$_{\pm1.57}$ & 16.68$_{\pm1.10}$ & 15.82$_{\pm2.98}$ & 18.02$_{\pm1.49}$ & 18.68$_{\pm0.89}$ & 15.81$_{\pm2.73}$ \\
\midrule
Mixtral & 1 & 18.71$_{\pm0.62}$ & 16.80$_{\pm1.20}$ & 14.67$_{\pm1.73}$ & 17.16$_{\pm1.81}$ & 16.58$_{\pm2.11}$ & 17.48$_{\pm0.79}$ & 17.31$_{\pm3.05}$ \\
Mixtral & 2 & 17.50$_{\pm1.98}$ & 16.12$_{\pm1.76}$ & 11.54$_{\pm3.66}$ & 13.74$_{\pm3.81}$ & 15.54$_{\pm3.48}$ & 16.54$_{\pm3.78}$ & 15.73$_{\pm3.32}$ \\
Mixtral & \textbf{3} & 18.86$_{\pm0.50}$ & 18.40$_{\pm0.86}$ & 16.29$_{\pm2.73}$ & 18.02$_{\pm1.86}$ & 18.86$_{\pm0.53}$ & 18.84$_{\pm0.51}$ & 18.06$_{\pm1.39}$ \\
Mixtral & 4 & 18.38$_{\pm0.70}$ & 17.48$_{\pm1.43}$ & 12.50$_{\pm2.39}$ & 16.22$_{\pm2.13}$ & 18.82$_{\pm0.56}$ & 18.80$_{\pm0.57}$ & 17.12$_{\pm1.97}$ \\
\midrule
Perplexity & 1 & 18.78$_{\pm0.46}$ & 17.32$_{\pm1.24}$ & 12.86$_{\pm1.48}$ & 17.28$_{\pm1.98}$ & 17.68$_{\pm0.91}$ & 18.08$_{\pm0.99}$ & 17.22$_{\pm3.75}$ \\ 
Perplexity & 2 & 16.84$_{\pm2.33}$ & 16.46$_{\pm2.07}$ & 14.14$_{\pm2.00}$ & 13.88$_{\pm2.97}$ & 16.00$_{\pm2.79}$ & 18.04$_{\pm2.18}$ & 14.52$_{\pm3.78}$ \\
 Perplexity & \textbf{3} & 19.00$_{\pm0.00}$ & 18.68$_{\pm0.47}$ & 17.40$_{\pm1.84}$ & 18.40$_{\pm1.03}$ & 19.00$_{\pm0.00}$ &    19.00$_{\pm0.00}$ & 18.53$_{\pm0.96}$ \\ Perplexity & 4 & 18.76$_{\pm0.69}$ & 17.20$_{\pm1.56}$ & 15.64$_{\pm1.75}$ & 15.68$_{\pm2.27}$ & 18.78$_{\pm0.91}$ & 18.76$_{\pm1.29}$ & 15.82$_{\pm3.53}$ \\
 \midrule
 Human &  & 17.65$_{\pm1.58}$ & 17.088$_{\pm1.72}$ & 16.69$_{\pm1.76}$ & 14.29$_{\pm3.23}$ & 17.96$_{\pm1.56}$ & 18.33$_{\pm1.03}$ & 15.29$_{\pm2.68}$ \\
 Yahoo \citet{chekalina-etal-2021-better} & - & 13.32$_{\pm4.97}$ & 9.32$_{\pm3.95}$ & 6.00$_{\pm1.00}$ & 5.21$_{\pm5.91}$ & 5.96$_{\pm3.64}$ & 5.96$_{\pm3.68}$ & 5.88$_{\pm4.86}$ \\
CAM \citet{shallouf-etal-2024-cam} & & 9.66$_{\pm4.43}$ & 9.52$_{\pm3.81}$ & 8.04$_{\pm2.78}$ & 6.54$_{\pm4.21}$ & 6.40$_{\pm2.29}$ & 6.46$_{\pm2.89}$ & 8.74$_{\pm4.56}$ \\
\end{tabular}}
    \caption{CompQA Benchmark leaderboard for all participating models, against each scenario.}
    \label{tab:results_all}
\end{table}

\end{document}